\documentclass[conference]{IEEEtran}
\IEEEoverridecommandlockouts
\usepackage{cite}
\usepackage{amsmath,amssymb,amsfonts}
\usepackage{algorithm, algpseudocode}
\usepackage{graphicx}
\usepackage{textcomp}
\usepackage{xcolor}
\usepackage{enumerate}
\usepackage{comment}
\usepackage{flushend}
\def\BibTeX{{\rm B\kern-.05em{\sc i\kern-.025em b}\kern-.08em
    T\kern-.1667em\lower.7ex\hbox{E}\kern-.125emX}}

\begin{document}

\title{A Study on Multi-Class Online Fuzzy Classifiers\\ for Dynamic Environments\\
}
% \author{\IEEEauthorblockN{Anonymous Authors}}
\author{
\IEEEauthorblockN{Kensuke AJIMOTO}
\IEEEauthorblockA{\textit{Graduate School of Informatics} \\
\textit{Osaka Metropolitan University}\\
Osaka, Japan \\
sc23962z@st.omu.ac.jp}
\and
\IEEEauthorblockN{Yuma YAMAMOTO}
\IEEEauthorblockA{\textit{Graduate School of Informatics} \\
\textit{Osaka Metropolitan University}\\
Osaka, Japan \\
sd24509i@st.omu.ac.jp}
\and
\IEEEauthorblockN{Yoshifumi KUSUNOKI}
\IEEEauthorblockA{\textit{Graduate School of Informatics} \\
\textit{Osaka Metropolitan University}\\
Osaka, Japan \\
yoshifumi.kusunoki@omu.ac.jp}
\and
\IEEEauthorblockN{\hspace{1.3cm}Tomoharu NAKASHIMA}
\IEEEauthorblockA{\hspace{1.3cm}\textit{Graduate School of Informatics} \\
\textit{\hspace{1.3cm}Osaka Metropolitan University}\\
\hspace{1.3cm}Osaka, Japan \\
\hspace{1.3cm}tomoharu.nakashima@omu.ac.jp
%ORCiD: 0000-0002-1443-0816
}
}

\maketitle

\begin{abstract}
This paper proposes a multi-class online fuzzy classifier for dynamic environments.
A fuzzy classifier comprises a set of fuzzy if-then rules where human users determine the antecedent fuzzy sets beforehand.
In contrast, the consequent real values are determined by learning from training data.
In an online framework, not all training dataset patterns are available beforehand.
Instead, only a few patterns are available at a time step, and the subsequent patterns become available at the following time steps.
The conventional online fuzzy classifier considered only two-class problems.
This paper investigates the extension to the conventional fuzzy classifiers for multi-class problems.
We evaluate the performance of the multi-class online fuzzy classifiers through numerical experiments on synthetic dynamic data and also several benchmark datasets.

% This document is a model and instructions for \LaTeX.
% This and the IEEEtran.cls file define the components of your paper [title, text, heads, etc.]. *CRITICAL: Do Not Use Symbols, Special Characters, Footnotes, 
% or Math in Paper Title or Abstract.
% This paper explores the performance of fuzzy classifiers under the context of multi-class and online learning. First, one-vs-one and one-vs-rest architectures are compared for a static online classification benchmark tests. Then, the performance of the fuzzy classifiers for dynamic classification problems are investigated where the data generation distributions changes over time.
\end{abstract}


\section{Introduction}
Machine learning, especially deep learning, has dramatically developed and shown its ability to achieve human-like performance in tasks such as image recognition and natural language processing.
In some tasks, such as board games, the machine learning model showed its overwhelming performance against human expert game players.
Although machine learning is highly able to learn from data, there is a claim that human users of a machine learning model cannot interpret its thinking process in a way humans can understand.

Fuzzy systems are advantageous in terms of their interpretability.
Unlike black-box models, such as neural networks, fuzzy systems are transparent and allow straightforward interpretation of their outputs.
This is because fuzzy systems model complex systems using human-understandable rules and linguistic variables.
These rules can be easily modified or updated by experts in the field, making fuzzy systems a powerful tool for decision-making and control systems.
Additionally, fuzzy systems can handle imprecise and uncertain data, making them suitable for applications where data is incomplete or noisy.
%↑ は本研究に特化した話ではなく，ファジィ全般のお話なので特に気にしなくてもよさそう
Overall, fuzzy systems offer a transparent and flexible approach to modeling complex systems, making them a valuable tool in various fields.

Meanwhile, fuzzy systems can be beneficial in a dynamic environment where data generation keeps moving over time because fuzzy systems can adapt to changing data and adjust their models in real time.
This kind of dynamical change in data generation is called concept drift in \cite{ConceptDrift}.
% データ生成源が時間とともに変化することは concept drift と呼ばれている[???]
For example, in a control system where the input patterns are constantly changing, a fuzzy system can adjust its rules to optimize the system's performance. 
Additionally, fuzzy systems can handle incomplete and noisy data, often in dynamic environments.
This ability allows for more robust and accurate models adapting to changing conditions. Overall, fuzzy systems are well-suited for dynamic environments because they adapt and handle uncertain data.

% このあたりで concept drift について言及できそう
% This paper proposes an online fuzzy classifier for dynamic enviroments that it is composed of fuzzy rules. The paper should include a description of its structure at the end of introduction section.

For two-class problems, a learning algorithm based on mathematical principles was proposed as a fuzzy classifier.
Although there are researches on fuzzy classifiers for multi-class problems such as the one in \cite{FuzzyBook}.

Multi-class problems with more than two classes have yet to be explored for the learning algorithm.
This paper considers how to handle multi-class problems by combining multiple binary fuzzy classifiers.

This paper is organized as follows: Section \ref{sec:BFC} describes a binary classifier along with its simple demonstration. How it interpretability is achieved is also explained. Next, multi-class classification by combining binary classifiers is addressed in Section \ref{sec:Multi-Classclassification}. Then Section \ref{NumericalExperiments} examines the classification performance of multi-class classification by fuzzy classifiers for both static and dynamic environments. Finally, Section \ref{Conclusions} summarizes this paper.

%\label{sec:Multi-Classclassification} ovr,ovo,online learning

\section{Binary Fuzzy Classifier}
\label{sec:BFC}
A binary fuzzy classifier considered in this paper is a classification model that takes a real-valued pattern as an input and gives a predicted class associated with the input.
The predicted class is either one of the prespecified two classes.
A fuzzy classifier consists of a set of fuzzy if-then rules.
There is an antecedent part and a consequent class in a fuzzy if-then rule.
The antecedent part is linguistically described, allowing human users to understand the covered area in the input space.

This paper considers an $n$-dimensional $C$-class classification problem.
We also assume that the input space is a unit two-dimensional space $[0, 1]^2$ without loss of generality.

\subsection{Fuzzy if-then rule}
We use the fuzzy if-then rule of the following style in this paper:

\begin{equation}
\label{eqn:fuzzy if-then rule}
\textrm{$R_j$: If $\vec{x}$ is $\vec{A}_j$ then $y=c_j$, $j=1,\ldots ,N,$}
\end{equation}
where $R_j$ is the label of the $j$-th fuzzy if-then rule, $\vec{x}=(x_1,\ldots , x_n)$ is an $n$-dimensional input pattern, $\vec{A}_j=(A_{j1},\ldots , A_{jn})$ is an $n$-dimensional antecedent fuzzy set and $c_j$ is a consequent real value that is equivalent to the support for a particular class (i.e., support for Class 1 or Class 2).

The $n$-dimensional antecedent fuzzy set is a conjunctive fuzzy set of $n$ one-dimensional fuzzy sets.
We use a product operator for the conjunction of multiple fuzzy sets.

\subsection{Classification of an unlabeled vector}
\begin{figure}[b]
%\vspace{5cm}
\centerline{\includegraphics[width=0.45\textwidth]{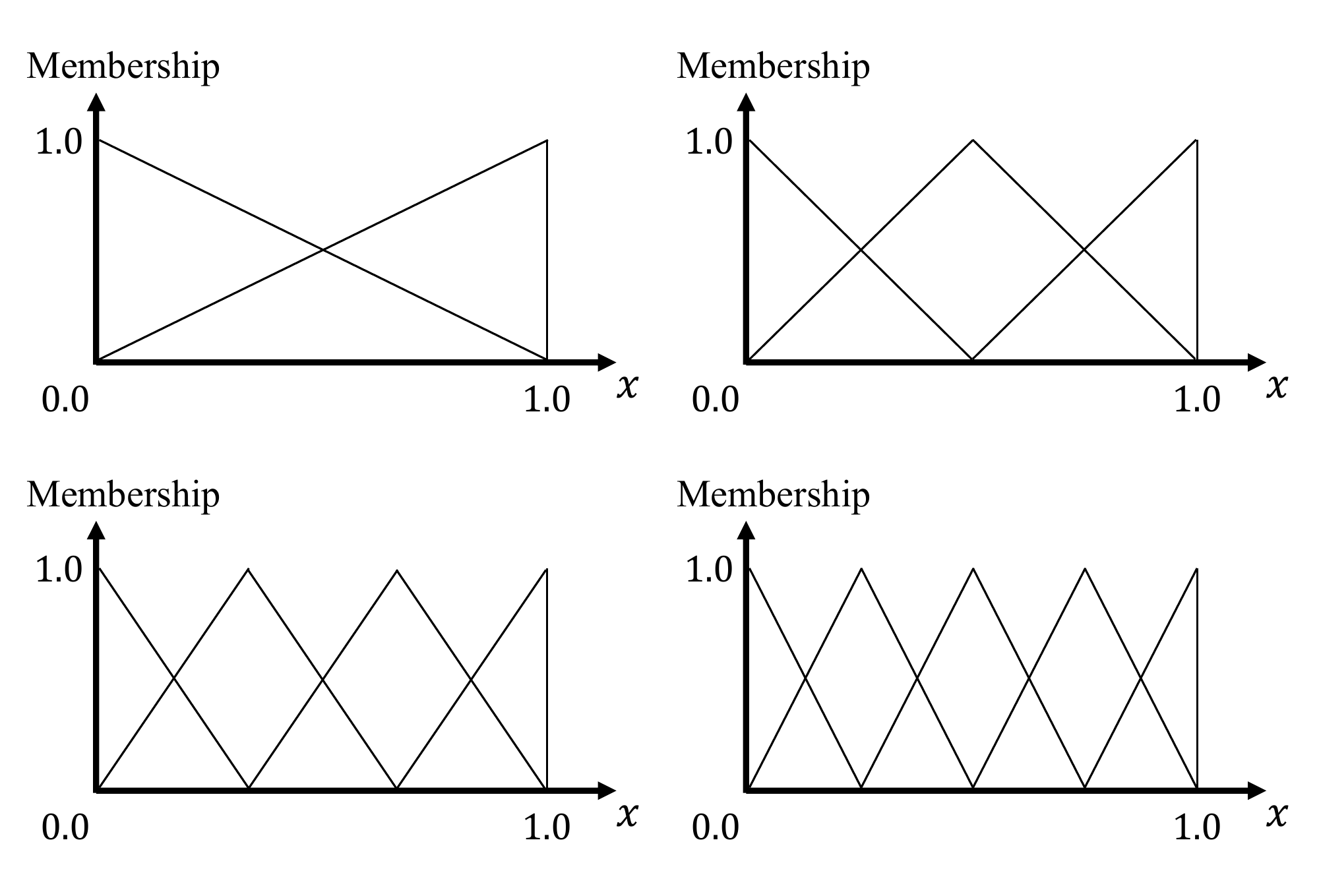}}
\caption{Triangular membership functions.}
\label{fig:triangular_mf}
\end{figure}

Assume we have constructed a fuzzy classifier with $N$ fuzzy if-then rules shown in the last subsection.
The class of an input pattern $\vec{x}$ is predicted by the fuzzy classifier using the following classification rule:

\begin{equation}
\label{eqn:classification rule}
\textrm{$\vec{x}$ is }
\left\{
    \begin{array}{ll}
        \textrm{Class 1,} & \textrm{if $\sum_{j=1}^N \mu_j(\vec{x})\cdot c_j \geq 0,$} \\
        \textrm{Class 2,} & \textrm{otherwise,}
    \end{array}
\right.
\end{equation}
where $\mu_j(\vec{x})$ is the membership value of the input pattern $\vec{x}$ for the $n$-dimensional fuzzy set $\vec{A}_j$ in $R_j$.
As the conjunction is defined by a product operator, $\mu_j(\vec{x})$ is calculated as follows:
\begin{equation}
\label{eqn:vec_A_j}
\mu_j(\vec{x})=\prod_{i=1}^n\mu_{ji}(x_i),
\end{equation}
where $\mu_{ji}(x_i)$ is the membership value of the $i$-th feature value for the fuzzy set $A_{ji}$.
$A_{ji}$ is a fuzzy set for the $i$-th feature and is assumed to be linguistically understandable such as \textit{small} and \textit{large}.
Human users can choose the shape of the membership function.
This paper uses a triangular shape for the membership function.
Fig. \ref{fig:triangular_mf} shows the triangular membership functions with various splittings of a feature.
\subsection{Interpretability}
\label{interpretability}
By definition, fuzzy if-then rules explain how to classify an input pattern.
We can interpret the classification process inside the fuzzy classifier because fuzzy rules contain linguistic fuzzy sets in their antecedent part.
However, when there are many fuzzy if-then rules in the fuzzy classifier, it is difficult for us to check all the fuzzy if-then rules.
In this case, picking up only a few fuzzy if-then rules important for classification is necessary to understand the process.

% 式\ref{eqn:classification rule} に注目すると，絶対値が大きい$c_j$は，クラス決定に及ぼす影響が大きい事がわかる．
% つまり$c_j$の値が最大であるファジィif-thenルール$R_j$はクラス1の代表的なルールと解釈できる．同様に$c_j$が負値で最も小さい値を取る場合，ルール$R_j$はクラス2の代表的なルールであると解釈できる．
% 
We can see from \eqref{eqn:classification rule} that fuzzy rules with a large absolute value of $c_j$ greatly influence the class decision.
In other words, the fuzzy if-then rule $R_j$ with the maximum value of $c_j$ are representative rules for their corresponding class.
If $c_j$ is positive, the fuzzy if-then rules contribute to Class 1.
Similarly, if $c_j$ is negative and takes the minimum value, the rule $R_j$ is representative of Class 2.

\subsection{Example of classification}
\begin{figure}[b]
\centering
\includegraphics[width=0.40\textwidth]{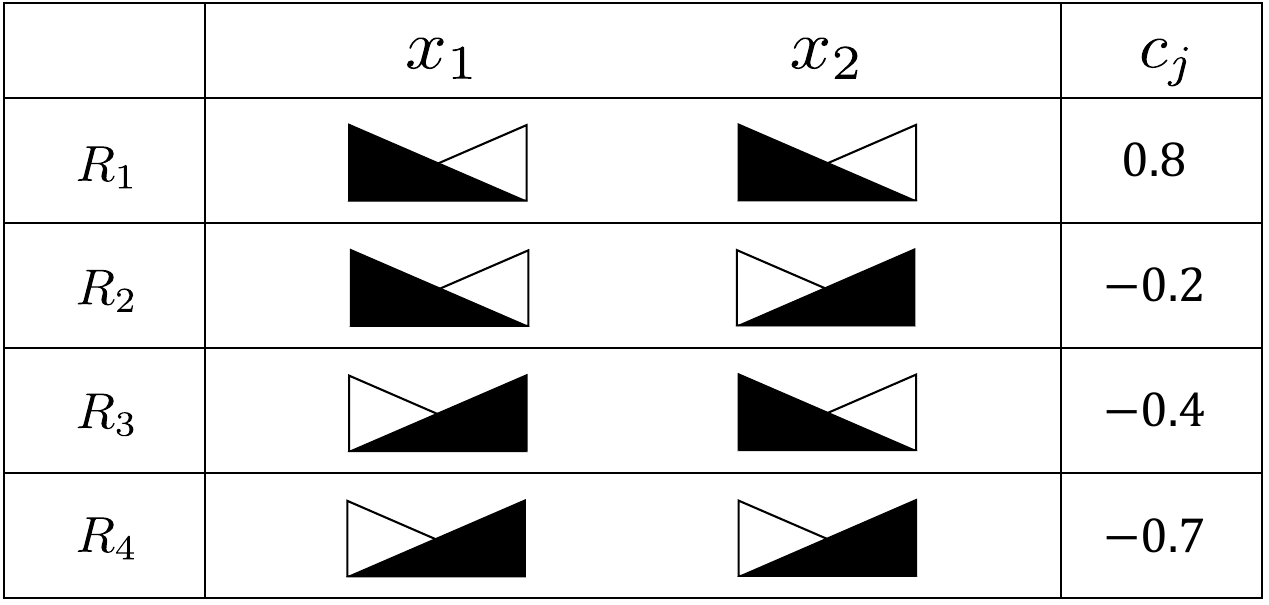}
\caption{Generated four fuzzy if-then rules in the example.}
\label{fig:example}
\end{figure}

This subsection describes an example of how to predict the class of an unlabeled input pattern.
Let us assume that we have generated four fuzzy if-then rules in \eqref{eqn:fuzzy if-then rule} for a two-dimensional classification problem.
The fuzzy sets shown in Table \ref{tab:settingOfFuzzyRule} are used for the antecedent fuzzy sets of fuzzy if-then rules.
In this example, we assign a random number with each of the antecedent fuzzy sets as shown in Fig.~\ref{fig:example} to generate four fuzzy if-then rules.
We show how to predict the class of an input pattern $\vec{x}=(x_1,x_2)=(0.2,0.4)$, which is unlabeled.
\begin{table}[b]
    \caption{An example of antecedent parts.}
    \label{tab:settingOfFuzzyRule}
    \centering
    \begin{tabular}{ccc}
      \hline 
       & $x_1$ & $x_2$\\
      \hline
      Rule 1 & \textit{small} & \textit{small} \\
      Rule 2 & \textit{small} & \textit{large} \\
      Rule 3 & \textit{large} & \textit{small} \\
      Rule 4 & \textit{large} & \textit{large} \\
      \hline
    \end{tabular}
\end{table}
In this case, the membership value of the input pattern for each fuzzy if-then rule is calculated as in Table \ref{tab:example}.
The predicted class of the input pattern is Class~1 according to the classification rule in \eqref{eqn:classification rule} because the total sum of the product between the membership value and the consequent value is positive.

\begin{figure}[tb]
%\vspace{5cm}
\centerline{\includegraphics[width=0.35\textwidth]{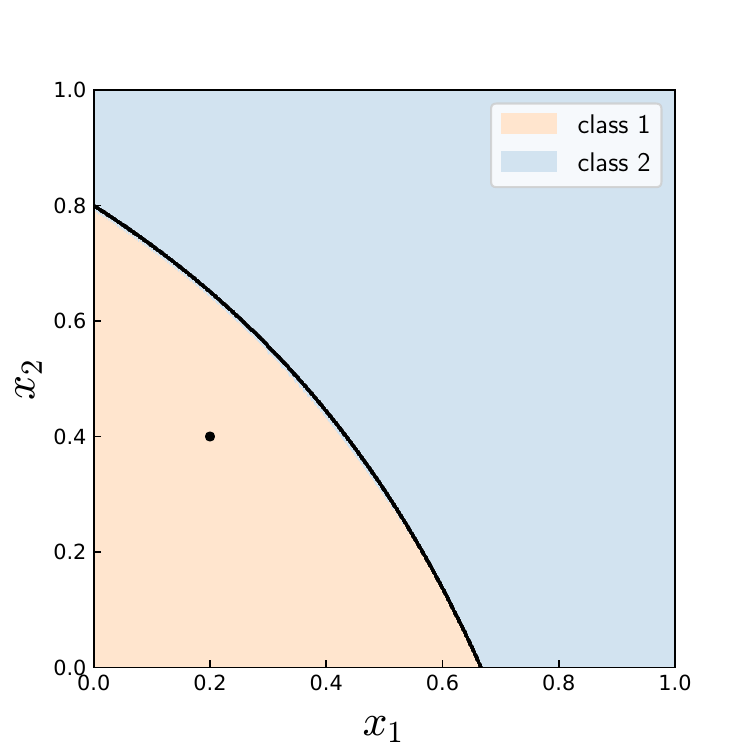}}
\caption{Generated boundary by the example fuzzy classifier.}
\label{fig:example boundary}
\end{figure}

Using the four fuzzy if-then rules in Table \ref{tab:example}, the classification boundary generated by the binary fuzzy classifier is drawn in the input space in Fig. \ref{fig:example boundary}.
From Fig. \ref{fig:example boundary}, we can see that the example input pattern $(0.2,0.4)$ falls in the area of Class~1.
\begin{table}[tb]
\caption{Calculation of membership values.}
\label{tab:example}
    \centering
    \begin{tabular}{cccccc}
        \hline
         $j$& $\mu_{j1}(x_1)$&  $\mu_{j2}(x_2)$&  $\mu_j(\vec{x})$& $c_j$ &$\mu_j(\vec{x})\cdot c_j$\\ \hline
         1&0.8&0.6&0.48&0.8&0.384\\
         2&0.8&0.4&0.32&-0.2&-0.064\\
         3&0.2&0.6&0.12&-0.4&-0.048\\
         4&0.2&0.4&0.08&-0.7&-0.056\\ \hline
         & &  & &  Total&0.216 \\
    \end{tabular}
\end{table}

\section{Multi-Class Online Classification}
\label{sec:Multi-Classclassification}
%二値分類器を組み合わせて多値分類に拡張するschemeとしてOne-vs-the-Rest(OvR)とOne-vs-One(OvO)が知られている．本章ではそれらについて説明を行う．
The fuzzy classifier in Section \ref{sec:BFC} is binary; Thus, it can be only applied to two-class problems.
This section describes how to use binary classifiers and accommodate multiple classes.
There are known schemes such as One-vs-the-Rest (OvR) and One-vs-One (OvO) that extend binary classifiers to multi-class classification\cite{MachineLearning:Peter,PatternRecognition:Bishop,Ensemble:Galar}.
This section explains these schemes.

\subsection{One-vs-the-Rest Scheme}
%OvRは$C$-class識別問題に対して$C$個の二値分類器を用いる．それぞれの識別器は，代表するクラスとそれ以外のクラスを分類する．つまり，1つ目の二値分類器は，入力がクラス1であれば正の値を，それ以外のクラスであれば負の値を出力する．すべての識別器の出力のうち最も大きなものに対応するクラスを識別器全体の識別結果とする．
The OvR scheme uses $C$ binary classifiers for a $C$-class problem.
Each of the $C$ classifiers focuses on only one class.
Thus, it distinguishes between the class it represents and all the other classes.
For instance, a binary classifier for Class 1, denoted as $f_1$, is expected to output a positive value for Class 1 input patterns and a negative value for the other classes.
The classifier with the highest output among all binary classifiers determines the overall classification result of input patterns.

\subsection{One-vs-One Scheme}
%OvOは$C$-class識別問題に対して$C(C-1)/2$個の二値分類問器を用いる．それぞれの識別器は異なる2つのクラスの組の間の分類を行う．つまり，1つ目の二値分類器は，入力がクラス1とクラス2のどちらであるかを識別する．また，本論文では最終的な識別結果は投票方式を採用している．したがって，最も多くの識別器から投票されたクラスが，識別器全体の識別結果となる．同率一位が発生した場合は，それらのうちからランダムなクラスを識別結果とする．また，学習データのラベルに関係しない二値分類器はそのデータを無視する．すなわち，先程の例での1つ目の二値分類器はクラス1とクラス2以外のラベル付けされた学習データを学習しない．
The OvO scheme uses $C(C-1)/2$ binary classifiers for a $C$-class problem.
Each classifier engages in the classification between only two different classes.
For example, a binary classifier for Class 1 and 2, denoted as $f_{1,2}$, uses only Class 1 and Class 2 data for training.
Thus, it sorts any input patterns into either Class 1 or Class 2 and never classifies them as any other classes.
The final classification result is determined by voting among the binary classifiers.
Therefore, the class that receives the most votes from the classifiers is the overall classification result.
If there is a tie, the result is a randomly selected class from those with the highest votes.

\subsection{Online Learning}
\label{sec:OnlineLearning}
Online learning\cite{OnlineLearning} is one of the crucial aspects of machine learning research that enables the adaptive development of models during training.
In contrast to common batch learning, where a model is trained on a fixed set of data, online learning allows models to learn and adapt to new data as it becomes available.
This is especially important in fields where data is constantly changing, such as finance, healthcare, and e-commerce.
Online learning also enables models to learn from their own mistakes, making them more accurate over time.
It can also improve efficiency by reducing the need for retraining models on large datasets, as updates can be made incrementally.
Overall, online learning plays a vital role in the development of machine learning models that can continuously improve and adapt to real-world data.
An online fuzzy classification system was proposed in \cite{evolving-fuzzy} where fuzzy rules are iteratively generated.
In \cite{CWFuzzyClassification}, Nakashima and Sumitani proposed online learning based on Confidence-Weighted learning \cite{CWlearning:Dredze} for the binary fuzzy classifier.
Although confidence-weighted learning is powerful, there is a disadvantage: it is not robust against noise.
This paper considers applying Passive-Aggressive (PA) \cite{PA:Crammer} learning instead.
PA is another online learning method that is robust against noise.
Wang et al.~\cite{PAFuzzyRobust:Wang} introduced PA learning in constructing a fuzzy system.
Their proposed method assigns a weight for each input pattern, and the weights are adjusted using PA.
This paper employs PA for adjusting the consequent weights in the similar manner as \cite{CWFuzzyClassification}.
We have conducted preliminary experiments where PA learning is compared with CW learning for a binary linear classifier.
As PA learning performed better than CW learning, we use PA learning in the experiments of this paper.

The classification rule in \eqref{eqn:classification rule} suggests that the linear combination of membership values determines the predicted class of an input pattern.
Thus, we can apply the Passive-Aggressive learning mechanism to the linear combination of the membership values.
Let us assume that the binary fuzzy classifier is to be trained using a training input pattern $\vec{x}$ from Class $y$ ($y=-1$ or $1$).
PA updates the consequent values (i.e., $\vec{c}=(c_1, \ldots ,c_N)$ in \eqref{eqn:classification rule}) 
with membership values $\vec{\mu}(\vec{x})=(\mu_1(\vec{x}),\ldots,\mu_N(\vec{x}))$ 
by the following equation:

\begin{equation}
    \vec{c}^{\textrm{~new}}=\vec{c}^{\textrm{~old}}
      +\frac{l(\vec{\mu}(\vec{x}),y,\vec{c}^{\textrm{~old}})}{\|\vec{\mu}(\vec{x})\|^2} \cdot
    y \cdot \vec{\mu}(\vec{x}),
    \label{eqn:PA}
\end{equation}
where $l(\cdot)$ is a hinge loss function defined as follows:
\begin{equation}
    l(\vec{\mu}(\vec{x}),y,\vec{c})=
    \begin{cases}
        0, & \textrm{if $y\cdot (\vec{c} \cdot \vec{\mu}(\vec{x})) \geq 1$},\\
        1-y\cdot (\vec{c} \cdot \vec{\mu}(\vec{x})), & \textrm{otherwise}.
    \end{cases}
    \label{eqn:hingeLoss}
\end{equation}

The overall procedure of the online learning is as in Algorithm \ref{arg:onlineLearning}.

\begin{algorithm}[b]
\caption{Online learning}
\label{arg:onlineLearning}
\begin{algorithmic}[1] % [1]を指定することで行番号が表示される
    \Require Dataset, Fuzzy partition of an input space
    \State $t \gets 1$
    \Loop
    \State Initialize a fuzzy classifier with $\vec{c}_t$
    \State Draw a labeled input pattern from a dataset
    \State Update the fuzzy classifier $\vec{c}_{t+1} \gets \vec{c}_t$ by \eqref{eqn:PA}
    \State $t \gets t+1$
    \EndLoop
\end{algorithmic}
\end{algorithm}

\section{Numerical Experiments}
\label{NumericalExperiments}
%本章では，提案手法の識別性能や解釈可能性を確認するために，データセットを用いたstatic problemに対する実験と，人工データを用いたdynamic problemに対する実験を行う．
In this section, we conduct experiments on static problems using some datasets and dynamic problems using synthetic data to verify the classification performance and interpretability of the proposed method. 

\subsection{Small number of antecedent fuzzy sets}
\label{Small number of antecedent fuzzy sets}
%Figure \ref{fig:triangular_mf} とは別に，Don't Care(DC)を定義する．DCは特徴軸の分割数が1である三角型メンバシップ関数であり，\ref{DC_set.pdf}のようになる．各軸を$m$分割するとすると，$m^n$個のルールが生成され，$n$が大きい場合，学習や識別に膨大な時間がかかる．そこで本研究では，1つ，もしくは2つの次元を選び，それらの特徴軸を$m$分割し，残りの次元はDCであるようなルールを用いて識別を行う．したがって，$n$次元問題に対し，$m^2\cdot{}_n C_2+m\cdot{}_n C_1+{}_n C_0$個のルールが生成される．
If each axis is divided into $m$ fuzzy sets, the number of generated fuzzy if-then rules is $m^n$.
Thus, the number of the generated fuzzy if-then rules exponentially increases as the number of dimensions increases.
The significant increase in fuzzy if-then rules leads to intractably long computation time.

\begin{figure}[tb]
\centerline{\includegraphics[width=0.3\textwidth]{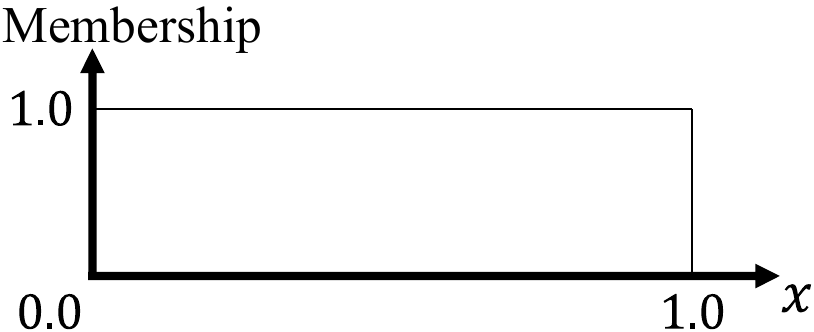}}
\caption{Don't Care membership function.}
\label{fig:DC}
\end{figure}

In order to remedy this problem, in addition to triangular membership functions in Fig.~\ref{fig:triangular_mf}, we introduce Don’t Care (DC).
The membership function of DC is rectangular where the output value is always $1$, as shown in Fig.~\ref{fig:DC}.

The numerical experiments in this paper limit the number of fuzzy sets used in a single fuzzy if-then rule to two at most.
In contrast, the other axes use DC instead; thus, these DC axes can be ignored in interpreting the fuzzy if-then rule.
% Therefore, for an $n$-dimensional problem, $m^2\cdot{}_n C_2+m\cdot{}_n C_1+{}_n C_0$ rules are generated, which is far smaller than $m^n$ rules.
%
% Cを使わない組み合わせの書き方に変更してみた．書式はもう少しきれいにできそう．
Therefore, for an $n$-dimensional problem, 
$
m^2\cdot
\big(\begin{smallmatrix}
  n\\
  2
\end{smallmatrix}\big)
+m\cdot
\big(\begin{smallmatrix}
  n\\
  1
\end{smallmatrix}\big)
+
\big(\begin{smallmatrix}
  n\\
  0
\end{smallmatrix}\big)
= m^2 \cdot \frac{n(n-1)}{2} + m \cdot n + 1
$
% ここまで追加
%
rules are generated, which is far smaller than $m^n$ rules.
It should be noted that for two-dimensional problems, we will not limit the number of fuzzy sets in the antecedent part of fuzzy if-then rules.
Thus, the number of fuzzy sets is always two in this case.

\subsection{Static Problems}
%本節ではいくつかのデータセットを用いてベンチマークテストを行う．用いたデータセットとそのcharacteristicsは\ref{tab:dataset}のとおりである．なお，学習や評価に用いる入力データは標準化している．
%fuzzy classifierは\ref{Small number of antecedent fuzzy sets}で述べた方式でファジィif-thenルールを生成し，各軸の分割数を$m=3$に設定している．fuzzy classifierの他に，2クラスPAとdelta learning (Widrow-Hoff Learning)をOvR,OvOを用いて多クラス識別可能にしたものを用いて比較を行う．PAとdelta learningはバイアス項を追加している．\ref{tab:resultForDataset}はstratified 10-fold cross validationを行い，すべての学習が終了した後のテストデータに対する正答率と学習にかかった時間の平均である．ただし，delta learningは学習率を決定する必要があるため，学習データのうち一部を学習率の決定のために用いている．
In this subsection, we perform benchmark tests using several datasets. Table \ref{tab:dataset} shows the datasets used and their characteristics.
Note that the input patterns are normalized in all datasets beforehand so that the input space is a two-dimensional unit space $[0, 1]^2$.
We generate fuzzy if-then rules as described in Subsection \ref{Small number of antecedent fuzzy sets}.
The number of partitions for each axis is set to $m=3$.
In addition to the fuzzy classifiers, we compare with PA (for non-fuzzy binary classification) and delta learning (non-fuzzy Widrow-Hoff learning) extended to multi-class classification using OvR and OvO.
PA and delta learning classifiers are both based on linear classifiers, and bias terms are employed in PA and delta learning to enhance classification performance.
Table \ref{tab:resultForDataset} shows the accuracy and required computational time for one epoch of online learning.
The accuracy refers to the correct classification rate on test data measured by stratified 10-fold cross-validation.
% stratified 10CV はよくわかってない
The time shows the computation time required for performing stratified 10-fold cross-validation.
Since delta learning requires determining the learning rate, a part of the training data is used for this purpose.
% ハイパーパラメータサーチのこと？

\subsubsection{Performance}
\label{static:performance}
%識別性能について，fuzzy classifierはOvR，OvOともにすべてのデータセットにおいて，delta learningよりも良い結果を示している．一方で，PAに対しては精度に大きな違いは見られなかった．また，fuzzy classifierのOvRとOvO間においても，そこまで差がなかった．
%学習時間について，fuzzy classifierは他アルゴリズムに比べて大きく劣る．これは特に次元数の大きな問題において顕著である．また，OvRとOvOを比較すると，OvOのほうが計算時間が短い．OvOのほうが識別器を構成するbinary classifierの数は多いものの，入力された学習データのクラスに関係しない識別器は学習をスキップするため，結果的に学習の回数はOvOの方が少ないためである．
In terms of classification performance, for both OvR and OvO, the fuzzy classifier shows better results than delta learning across all datasets. However, there was no significant difference in accuracy compared to PA. Also, there was not much difference between OvR and OvO for the fuzzy classifier.

Regarding the learning time, the fuzzy classifier is significantly inferior to other algorithms, especially in problems with high dimensionality. However, by parallelizing the calculation of the fuzzy classifier’s membership values, there is a possibility to shorten the learning time. Also, when comparing OvR and OvO, OvO has a shorter computation time. Although OvO has more binary classifiers constituting the classifier, it skips learning for classifiers that are irrelevant to the class of the input training patterns, resulting in a reduction in the number of learnings for OvO.
The speedup of computational time is not the focus of this paper this is left for future research.

\subsubsection{Interpretability}
\label{static:interpretability}

Here, we show the interpretation of the resultant fuzzy classifier by observing the consequent real value $c$ of the fuzzy classifier after training on the iris dataset.
In one of the 10-fold cross-validations, we focus on one of the OvR classifiers, $f_{versicolor}$. The rule with the largest $c$ was \textit{If petal length is medium and petal width is medium}, and the rule with the smallest was \textit{If petal width is large}. As shown in section \ref{interpretability}, rules with large absolute values of $c$ highly contribute to the class determination. Therefore, it can be interpreted that $f_{versicolor}$ learned to emphasize these rules.

\begin{figure}[tb]
%\vspace{5cm}
\centerline{\includegraphics[width=0.45\textwidth]{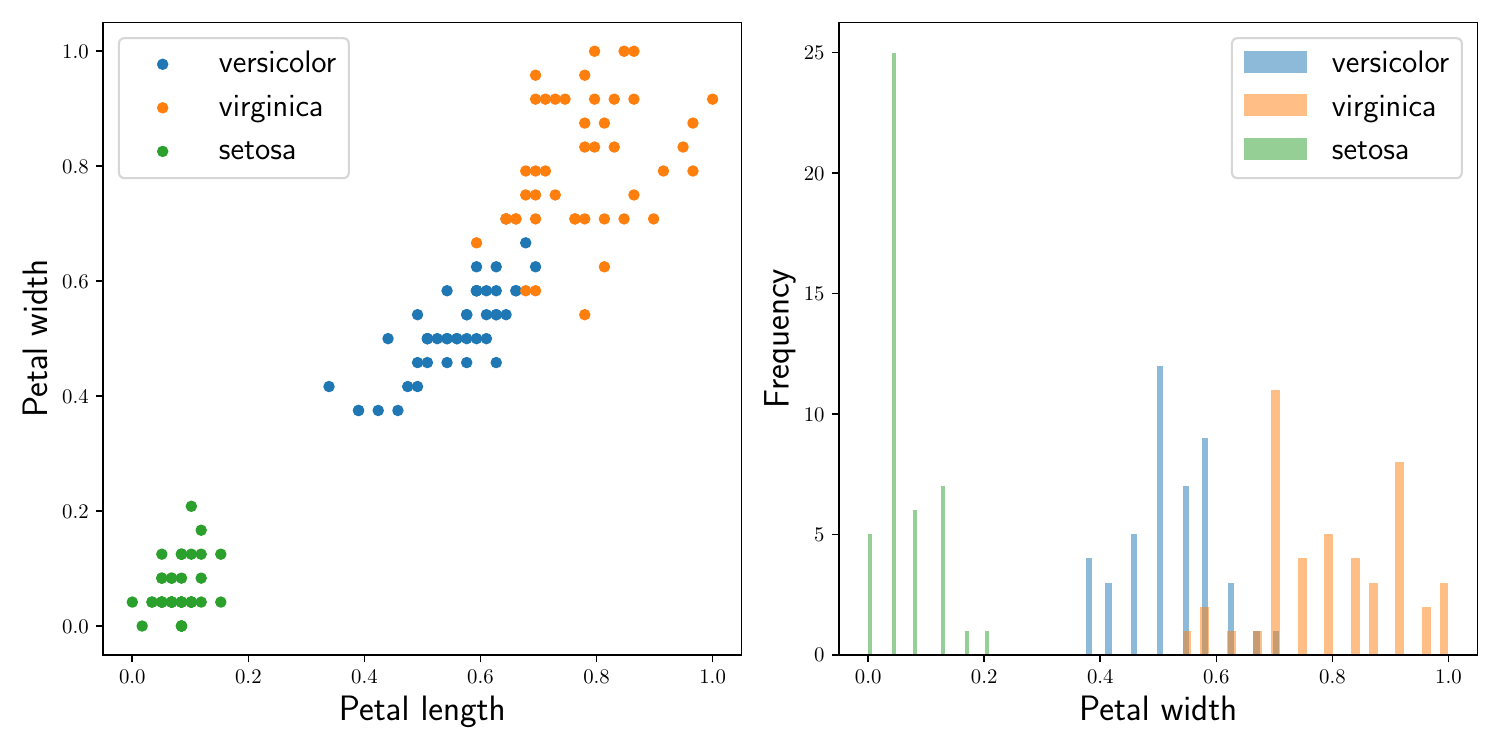}}
\caption{Distributions of petal length and petal width of the iris dataset.}
\label{fig:irisPlot}
\end{figure}

The scatter plot matrix corresponding to these rules is shown in Fig.~\ref{fig:irisPlot}. The region of the rule with a large value of $c$ had a lot of versicolor data, and the region of the rule with a small value had a lot of virginica. This trend was the same for the top 5 rules with large and small values.
Hence, $f_{versicolor}$, which distinguishes versicolor from the other classes, contributes greatly to the identification of non-versicolor classes, especially the rules of the region where virginica existed. The distribution of the virginica data matches the area covered by a rule with a large consequent value. Therefore, it is considered that the value of $c$ of the rules in the region where virginica data exist is particularly small so that virginica can be correctly identified (i.e., as “the rest”).
Similarly, in OvO, it is also possible to check which rules are prioritized in distinguishing between classes $a$ and $b$ using the consequent real value of the binary classifier $f_{a,b}$.

\subsubsection{Summary}
%識別性能と説明可能性はトレードオフの関係にある\cite{FuzzyBook}．一方で，fuzzy classifierはdelta learningよりも高い識別性能を持ち，またPAと同程度の精度で識別が可能であることが確認された．また，ファジィルールとその後件部実数を用いて，モデルの解釈が可能である．つまりfuzzy classifierは識別性能を保ちつつ，説明可能性を獲得したモデルであると言える．
Classification performance and interpretability are in a trade-off relationship \cite{FuzzyBook}. 
It has been confirmed from the experimental results that the fuzzy classifier has a classification performance superior to delta learning and comparable to PA.
Nevertheless, by using the fuzzy rules and their consequent real value, it is possible to interpret the model.
In other words, the fuzzy classifier is a model that has acquired interpretability while maintaining good classification performance.

\begin{table}[b!]
\caption{Dataset characteristics}
\label{tab:dataset}
    \centering
    \begin{tabular}{lccc}
        \hline
        Dataset&Instance&Class&Dim\\
        \hline
        Speaker accent recognition\cite{UCI}&5473&5&10\\
            Crabs\cite{Crab}&200&5&4\\
        Glass identification\cite{UCI}&214&7&9\\
        Iris\cite{UCI}&150&3&4\\
        Page blocks classification\cite{UCI}&5473&5&10\\
        Seeds\cite{UCI}&210&3&7\\
        Wine\cite{UCI}&178&3&13\\
    \hline
    \end{tabular}
\end{table}

\begin{table*}[t]
\caption{Performance comparison of Fuzzy classifier, PA, and Delta learning on benchmark datasets}
    \centering
    \begin{tabular}{lcccccc}
        \hline
        dataset&Fuzzy(OvR)&Fuzzy(OvO)&PA(OvR)&PA(OvO)&Delta(OvR)&Delta(OvO)\\
        \hline
        \textbf{Speaker accent recognition}&&&\\
        \hspace{6pt}accuracy (\%)&$\mathbf{52.36}\pm10.98$&$50.80\pm13.88$&$41.63\pm15.96$&$41.63\pm16.58$&$25.76\pm7.32$&$45.15\pm3.33$\\
        \hspace{6pt}time (s)&$314.82\pm1.68$&$267.50\pm0.97$&$1.94\pm0.71$&$\mathbf{1.67}\pm0.82$&$2.47\pm0.87$&$2.44\pm0.92$\\
        \textbf{Crabs}\\
        \hspace{6pt}accuracy (\%)&$\mathbf{45.29}\pm10.09$&$40.66\pm6.16$&$28.66\pm4.82$&$32.61\pm6.56$&$25.00\pm0.00$&$29.5\pm4.38$\\
        \hspace{6pt}time (s)&$12.90\pm1.21$&$9.96\pm1.24$&$0.76\pm0.67$&$\mathbf{0.68}\pm0.67$&$1.04\pm0.67$&$1.04\pm0.70$\\
        \textbf{Glass identification}\\
        \hspace{6pt}accuracy (\%)&$53.70\pm11.31$&$\mathbf{57.40}\pm10.84$&$44.89\pm3.52$&$46.73\pm4.85$&$27.73\pm6.93$&$43.27\pm2.20$\\
        \hspace{6pt}time (s)&$109.76\pm0.77$&$96.90\pm1.67$&$1.32\pm0.68$&$\mathbf{1.27}\pm0.71$&$1.85\pm0.66$&$1.71\pm0.81$\\
        \textbf{Iris}\\
        \hspace{6pt}accuracy (\%)&$\mathbf{96.67}\pm4.71$&$\mathbf{96.67}\pm4.71$&$82.67\pm7.83$&$92.67\pm5.84$&$70.67\pm8.43$&$89.33\pm12.25$\\
        \hspace{6pt}time (s)&$4.17\pm0.84$&$2.85\pm0.80$&$0.49\pm0.66$&$\mathbf{0.41}\pm0.65$&$0.59\pm0.67$&$0.47\pm0.67$\\
        \textbf{Page blocks classification}\\
        \hspace{6pt}accuracy (\%)&$92.93\pm0.57$&$\mathbf{93.38}\pm0.40$&$92.09\pm0.51$&$91.94\pm0.86$&$87.90\pm2.77$&$81.26\pm3.02$\\
        \hspace{6pt}time (s)&$2471.10\pm3.99$&$1977.97\pm9.46$&$12.42\pm0.77$&$\mathbf{10.37}\pm0.81$&$20.46\pm0.68$&$17.13\pm0.76$\\
        \textbf{Seeds}\\
        \hspace{6pt}accuracy (\%)&$89.48\pm8.03$&$88.05\pm8.75$&$91.38\pm8.04$&$\mathbf{91.88}\pm7.78$&$75.71\pm7.92$&$82.86\pm16.06$\\
        \hspace{6pt}time (s)&$23.98\pm1.17$&$16.01\pm1.04$&$0.68\pm0.65$&$\mathbf{0.55}\pm0.66$&$0.92\pm0.68$&$0.67\pm0.70$\\
        \textbf{Wine}\\
        \hspace{6pt}accuracy (\%)&$\mathbf{97.75\pm2.91}$&$96.63\pm3.91$&$96.67\pm7.03$&$92.12\pm7.52$&$92.78\pm10.49$&$72.78\pm9.96$\\
        \hspace{6pt}time (s)&$105.90\pm1.49$&$71.34\pm1.31$&$0.90\pm0.66$&$\mathbf{0.79}\pm0.67$&$1.16\pm0.69$&$0.90\pm0.71$\\
        \hline
    \end{tabular}

\label{tab:resultForDataset}
\end{table*}
%
\begin{comment}
\begin{table}[h]
\caption{Top 4 rules with the highest consequent real value $c$}
\label{tab:highestWeightForIris}
    \centering
    \begin{tabular}{l@{\hskip 5pt}l@{\hskip 3pt}l@{\hskip 3pt}c@{\hskip 5pt}l@{\hskip 5pt}l@{\hskip 3pt}l@{\hskip 3pt}c|r}
        \hline
        \multicolumn{8}{c}{Antecedent part} & \multicolumn{1}{|c}{$c$}\\
        \hline
        If&petal length&is& medium&and&petal width&is&medium&$0.90$\\
        If&petal width &is& medium&&&&&$0.68$\\
        If&sepal width &is& medium&and&petal width&is&medium&$0.56$\\
        If&petal length&is& medium&&&&&$0.56$\\
        %If&sepal width &is& medium&and&petal length&is&medium&$0.44$\\
    \hline
    \end{tabular}
\end{table}
%
\begin{table}[h]
\caption{Top 4 rules with the lowest consequent real value $c$}
\label{tab:lowestWeightForIris}
    \centering
    \begin{tabular}{l@{\hskip 5pt}l@{\hskip 3pt}l@{\hskip 3pt}c@{\hskip 5pt}l@{\hskip 5pt}l@{\hskip 3pt}l@{\hskip 3pt}c|r}
    \hline
        \multicolumn{8}{c}{Antecedent part} & \multicolumn{1}{|c}{$c$}\\
    \hline
        If&petal width&is&large&&&&&$-0.72$\\
        If&petal length&is&large&&&&&$-0.58$\\
        If&sepal length&is&medium&and&petal width&is&large&$-0.57$\\
        If&sepal width &is&medium&and&petal width&is&large&$-0.51$\\
        %IF&sepal length&is&medium&and&petal length&is&large&$-0.47$\\
    \hline
    \end{tabular}
\end{table}
\end{comment}
%
\subsection{Dynamic Problems}
\begin{figure}[htbp]
\centerline{\includegraphics[width=0.35\textwidth]{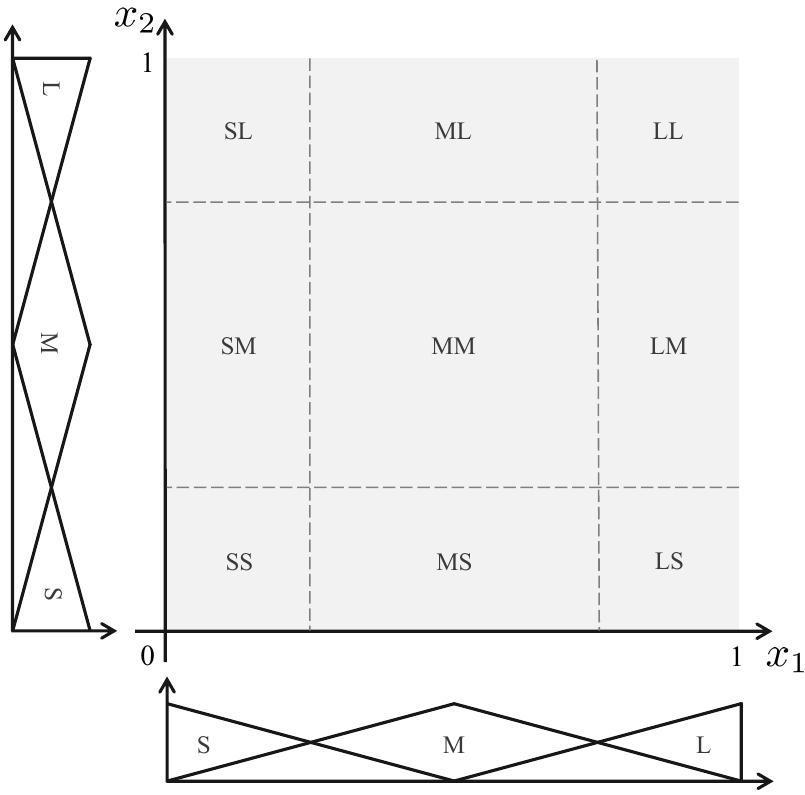}}
\caption{An overview of the area covered by the rules generated from the three-partition membership functions.}
\label{fig:3x3_partition}
\end{figure}

% この節で言いたいこと
% ・多クラスに拡張したPA学習ファジィ識別器が動的に変化する問題に追従できるかを確認
% ・学習器の重みの変化を確認することで各時刻で識別器が重要視しているルールを確認
% ・One vs the Rest では各識別器が 1 つのクラスとその他のクラスを識別するため，任意の識別器を調査することで，それぞれのクラスに対する挙動を確認することができる．
% ・One vs One では各識別器が任意の 2 クラスを識別するため，識別器を調査することで任意の 2 クラスの関係を確認することができる
%
% 動的問題の実験では，提案手法(fuzzy classifier)が時間経過とともに変化するような問題環境に対応できるかどうかを確かめるため，人工データを用いて数値実験を行う．
% \ref{sec:Multi-Classclassification}章で述べたOvR,OvOのそれぞれの戦略を用いてfuzzy classifierを多クラスに拡張し，問題環境が動的に変化する2次元3クラス分類問題を学習させる．
% 以下，この問題を動的オンラインパターン識別問題とよぶ．
% 実験の詳細を述べる．
For dynamic problems, numerical experiments are conducted on synthetic data to see if the proposed method can handle problem environments that change over time.
Using each of the OvR and the OvO schemes described in section \ref{sec:Multi-Classclassification}, we extend the fuzzy classifier to multi-class classification to train a two-dimensional, three-class problem in which the problem environment dynamically changes over time.
% Hereafter, this problem will be referred to as the dynamic online pattern classification problem.
% すべての識別器で使用するメンバシップ関数はFig. \ref{fig:triangular_mf} で示された3分割のものとする．
% \ref{fig:3x3_partition}に生成されるルールが2次元平面上で覆う領域のイメージを示す．
We use three fuzzy sets for each axis for all classifiers in this subsection.
The membership functions used in the experiments of this section are triangular as shown in Fig.~\ref{fig:triangular_mf}.
An overview of the area covered in two-dimensional space by the rules generated from the three membership functions for each axis is shown in Fig. \ref{fig:3x3_partition}.
%
% 式\ref{eqn:classification rule} より，$\vec{c_j}$の値が最大となるルール$R_j$は識別において最も重要視するルールと解釈できる．
% 以上を踏まえて，各識別器が各時刻で持つ最大の$c$の変化を観察することで正規分布の移動に追従して学習できているかを調査する．
% 各クラスはそれぞれ独立した多変量正規分布$\mathcal{N}$$(\vec{\mu}_c, \vec{\sigma}_c)$，$c=1,\ldots ,C$を保持する．ここで，$C$はクラス数である．生成する学習用パターンのクラスラベルは毎回ランダムに選択され(クラスの出現比率を変更することもできる)，学習用パターンの特徴量は選択されたクラスcの多変量正規分布$\mathcal{N}$$(\vec{\mu}_c, \vec{\sigma}_c)$に従う．
% また，正規分布のパラメータは次のとおりである．
\begin{comment}
From \eqref{eqn:classification rule}, the rule $R_j$ with the largest consequent real value $c_j$ can be interpreted as the most important rule in the classification.
Considering the above, this experiment investigates whether the classifiers are able to learn by following the movement of the Gaussian distributions by observing the change in the maximum $c_j$ each classifier has at each time.
\end{comment}
As described in Subsection \ref{interpretability}, the fuzzy if-then rules with a high consequent real value are essential in the classification.
So, they must be used for the interpretation of the classification process.
Considering this, this subsection investigates whether the classifiers can learn under a dynamic environment by following the transition of class distributions while adapting the rule.
We also want to observe whether the most important rule (i.e., the rule with the maximum consequent real value) for each class follows the transition of the class distributions.
Three classes follows a two-dimensional Gaussian with a mean $(\frac{2-\sqrt{2}}{4},\frac{2-\sqrt{2}}{4})$, $(0.5,0.5)$, $(\frac{2+\sqrt{2}}{4},\frac{2+\sqrt{2}}{4})$, respectively, and a diagonal covariance matrix with $(0.1, 0.1)$ on its diagonal.
% Each class has a two-dimensional Gaussian with mean $\vec{m}_l, (l=1,2,3)$ and a diagonal covariance matrix with $(0.1, 0.1)$ on its diagonal.
% Mean vectors of the Gaussians are difined as follows:
% %
% \begin{align}
%     \vec{m}_1 &= (\frac{2-\sqrt{2}}{4},\frac{2-\sqrt{2}}{4})^\top,\\
%     \vec{m}_2 &= (0.5, 0.5)^\top,\\
%     \vec{m}_3 &= (\frac{2+\sqrt{2}}{4},\frac{2+\sqrt{2}}{4})^\top.
% \end{align}
%

\begin{figure}[tb]
\centerline{\includegraphics[width=0.4\textwidth]{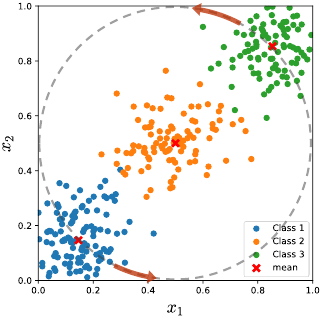}}
\caption{The initial setup of three classes for a dynamic problem.}
\label{fig:imageOfGaussian}
\end{figure}

During the experiments, training patterns are generated randomly from one of the three Gaussian distributions.
The class label of the training pattern is first randomly picked up every time a training pattern is generated (i.e., the class appearance ratio is equal).
The input pattern of the training pattern is generated using the Gaussian of the selected class.
% クラス1とクラス3の平均ベクトルのパラメータは，それぞれが回転移動したときに定義式$[0,1]^2$から外れないように定めた．具体的には回転中心と分布の平均ベクトルの距離がが0.5に近い値で収まるように設定した．
% また，定義域の外にパターンが生成された場合は，定義域内に収まるまで生成し直すこととする．
The class distributions move during the experiments to make the problem dynamic.
The initial mean vectors for Class 1 and Class 3 are set so that the positions of the two classes are symmetric relative to the rotation center $(0.5, 0.5)$.
We carefully consider the position of the means so that they do not go outside the unit space $[0, 1]^2$, as the unit space is the domain in this experiment.
% The values of $\vec{m}_1$ and $\vec{m}_3$ should be set so that the distance from the point $(0.5,0.5)$ to $\vec{m}_l$ and $\vec{m}_3$ respectively is within a value close to 0.5.
% それぞれの分布から生成されるパターンの例を\ref{fig:imageOfGaussian}に示す．
Fig. \ref{fig:imageOfGaussian} shows the initial setup of the means of Class 1 and Class 3.
We also show the rotation direction of Class 1 and Class 3 by two red curved arrows.
It should be noted that the points of Class 1--3 are only to show the concept of the dynamic environment.
In the actual experiments, we incrementally generate only a few training patterns from the corresponding Gaussian distributions, and the training patterns are abandoned right after they are used for training fuzzy classifiers:
The generated training patterns are never used again for training.

Numerical experiments are conducted as follows:
% step 1. 学習用パターンを決められた数だけその時刻における正規分布から生成．
% step 2. 生成されたパターンを1つずつ学習
% step 3. 任意の数のパターンがすべて与えられた後，正規分布の平均ベクトルをあらかじめ指定された中心点に対して反時計回りに1度回転移動
\begin{enumerate}[Step. 1]
    \item Generate a prespecified number of training patterns from the current Gaussian distributions.
    \item Training on the generated patterns.
    \item After all patterns are given, all mean points of all classes are rotated $1$ degree counterclockwise around a prespecified center point.
\end{enumerate}
% 上記の手順を分布が360度回転 (1周) するまで繰り返す．
The procedure described above is repeated until the point when the distribution of each class has rotated 360 degrees.
% 一時刻で各クラスの分布を十分に学習させるため，step1.で生成される学習用パターンの数を10とした．
In this experiment, the number of training patterns generated in Step 1. is set to 10.
% In this experiment, the number of training patterns generated in Step 1. is set to 10 in order to sufficiently learn the distribution of each class at each moment in time.
% したがって，この実験で生成される学習用パターンの総数は3600個である．
% また，step 3. における回転中心を$(0.5,0.5)$とする．
% % 式\ref{eqn:classification rule} より，$\vec{c_j}$の値が最大となるルール$R_j$は識別において最も重要視するルールと解釈できる．
% 以上を踏まえて，各識別器が各時刻で持つ最大の$c$の変化を観察することで正規分布の移動に追従して学習できているかを調査する．
Thus, the total number of training patterns generated is 3600.
The center of rotation in Step 3. is set to $(0.5,0.5)$.
% From \eqref{eqn:classification rule}, the rule $R_j$ with the largest consequent real value $c$ can be interpreted as the most important rule in the classification.
Considering the above, this experiment investigates whether the classifiers are able to learn by following the movement of the Gaussian distributions by observing the change in the maximum $c$ each classifier has at each time.
\subsubsection{OvR scheme}
\begin{figure}[tb]
\centerline{\includegraphics[width=0.45\textwidth]{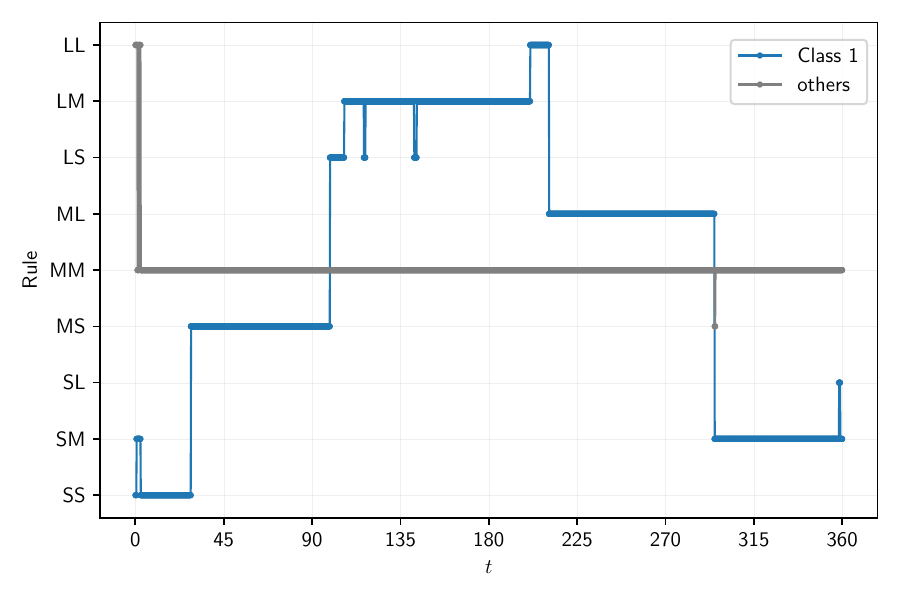}}
\caption{Transitions of the important rules in $f_1$.}
\label{fig:OvR_3600_clf1}
\end{figure}

% OvR scheme で学習した場合のClass 1とその他のクラスを識別する識別器の最大また最小はの$c_j$を持つルール$R_j$の変化を\ref{fig:OvR weight changes}に示す．
% グラフの横軸$t$は時刻を，縦軸はルールを表す．
% 青のラインは最大の後件部実数値$c$をもつルールを，グレーのラインは最小の$c$をもつルールを示している．
% 青のラインに注目すると，SSからスタートし最後にSMにたどり着くまで\ref{fig:3x3_partition}の反時計回りに外周を1周するような遷移をすることが読み取れる．
% 例えば，$t=45$のとき，Class1の分布の中心は反時計回りに$45$度回転するため，Fig.\ref{fig:3x3_partition.pdf} で示すMSの領域に位置する．また，Fig.\ref{fig:OvR weight changes}から$c$が最大となるルールはMSであることがわかる．
The changes in rule with the maximum and minimum consequent real value $c$ of $f_1$ when trained with the OvR scheme are shown in Fig.~\ref{fig:OvR_3600_clf1}.
The horizontal axis $t$ of the graph represents the time and the vertical axis the rule. The blue line shows the rule with the maximum $c$ and the grey line shows the rule with the minimum $c$.
Focusing on the blue line, it can be read that the transition starts from rule SS and circles around the outer circumference counterclockwise in Fig.~\ref{fig:OvR_3600_clf1} until it finally reaches rule SM.
When $t=45$, for example, the center of Class 1 distribution rotates $45$ degrees counterclockwise, and therefore its mean is located in the area of rule MS shown in Fig.~\ref{fig:3x3_partition}. From Fig.~\ref{fig:OvR_3600_clf1}, it can also be seen that the rule with the maximum $c$ at $t=45$ is rule MS.
% このように，識別器の$c$の値を調べることでその時刻において識別器が重要視しているルールを知ることができる．
In this way, by examining the consequent real value $c$ of the classifier, it is possible to reveal the rule that the classifier attaches importance to at that time.
% $f_1$, $f_2$, $f_3$, $f_{1,2}$, $f_{1,3}$, $f_{2,3}$
%
\subsubsection{OvO scheme}
\begin{figure}[tb]
\centerline{\includegraphics[width=0.45\textwidth]{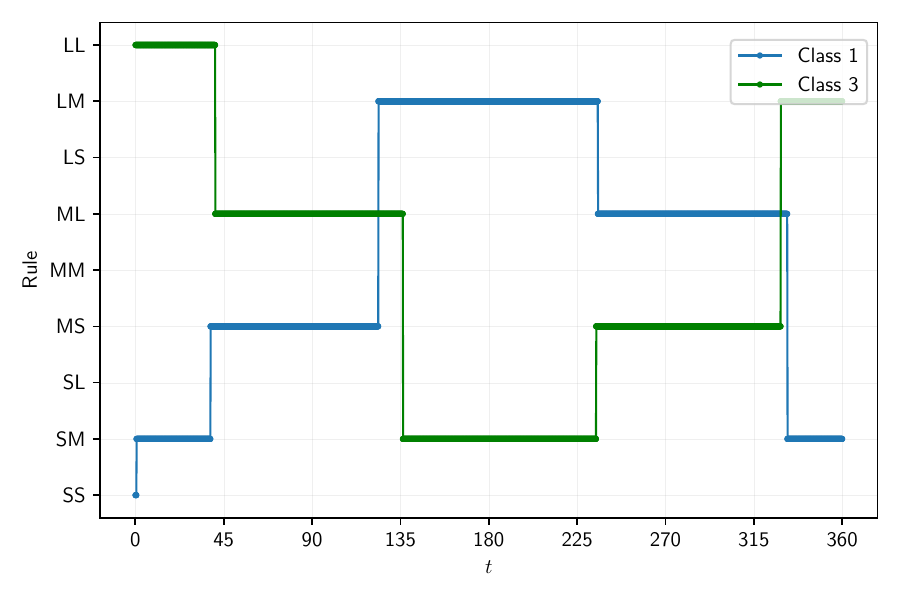}}
\caption{Transitions of the important rules in $f_{1,3}$.}
\label{fig:OvO_3600_clf1vs3}
\end{figure}

% 一例として，クラス1とクラス3を識別する識別器の$c_j$の変化を確認する．
% OvRと同様にFig.~\ref{fig:OvO_3600_clf1vs3}は各時間における最大$c_j$と最小$c_j$の変化を示している。
% 青い線がクラス1、緑の線がクラス3の代表的ルールを表す．
% ここで示す識別器は\ref{eqn:classification rule} におけるClass 2 をClass 3に置き換えたものである．
% Fig.~\ref{fig:OvO_3600_clf1vs3}から，Class 1, Class 3 はともに\ref{fig:3x3_partition}の外周を1周するような遷移をしていることが読み取れるため，分布の移動に追従して学習できている事がわかる．
% また，残りの2つのクラスの組み合わせでも同様に分布の変化を捉えていることが確認できた．
As an example, check the change in the consequent real value $c$ of the $f_{1,3}$.
Similar to OvR, Fig.~\ref{fig:OvO_3600_clf1vs3} shows the maximum and minimum change in $c$ at each time.
The blue line indicates Class 1 and the green line indicates Class 3.
The classifier shown here replaces Class 2 in \eqref{eqn:classification rule} with Class 3.
From Fig.~\ref{fig:OvO_3600_clf1vs3}, it can be observed that both Class 1 and 3 transition around the outer circumference of Fig.~\ref{fig:3x3_partition}, which indicates that the classifier is able to learn following the movement of the distributions.
It was confirmed that the remaining $f_{1,2}$ and $f_{2,3}$ could also learn to follow changes in the distributions in the same way.
\subsubsection{Summary}
% 実験のまとめ
% 精度はどちらも97.5%だった.
% また，どちらのschemeにおいても学習中に識別器の重要視しているルールを明らかにできることを示した．
% しかし，$110 \leq t \leq 150$などで見られるように、識別器が過去に平均が位置していた領域を依然重視している時刻がある．
% これにより，解釈性が【低下】する恐れがある．
% 過去の学習影響を漸減させるために少しずつcの値を0に近づける（過去を忘れる）ことでこの懸念は解決できる可能性があると考える．
% この問題は今のところオープンプロブレムthese今後解決されるべき問題である．
In these experiments, the classification accuracy is about 97.5\% for both OvR and OvO.
We also showed that in both schemes it is possible to reveal the rules that are important to the classifier during learning.
However, there are also times when the classifier still focuses on the area where the mean was located in the past, as seen for example in $110 \leq t \leq 150$. This may reduce interpretability.
We suggest that this concern could be addressed by gradually adjusting the consequent real values $c$ closer to $0$ (i.e. forgetting the past) in order to reduce the learning influences of the past.

\section{Conclusions}
\label{Conclusions}
%本論文では，オンライン学習アルゴリズムであるPAを導入したファジィ識別器を，OvR，OvO schemeを用いて多クラスに拡張した．
%実験
%いくつかのベンチマークデータセットを用いたstatic problemに対する実験では提案手法は，OvRやOvOを用いた多クラス識別PAと同等の識別性能を発揮することが確認された．
% 人工データを用いたdynamic problemに対する実験では，時刻とともに移動する複数の分布を学習できることを示した．
% また，それぞれの実験で，ファジィif-thenルールを確認することで識別器の解釈性も維持されていることがわかった．
% 今後の展望として，動的問題では，過去の学習を忘却することで解釈性を向上させることが可能であるかの調査が挙げられる．
% 
In this paper, a binary online learning fuzzy classifier was extended to multi-class classification by using OvR and OvO schemes.
Experiments on the static problem using several benchmark datasets confirmed that the proposed method achieves comparable classification performance to multi-class Passive-Aggressive classifiers using OvR and OvO.
Experiments on the dynamic problem using synthetic data confirmed that the proposed method is able to learn multiple distributions that move over time and follow their changes.
In each experiment, it was also shown that the classifier can be interpreted by examining the fuzzy if-then rules.
During the experiments, we observed some perturbation of the important rules.
This was caused because the learning results of the fuzzy if-then rules remains even if they are obsolete after changing the class distribution moves away from the covered area by the fuzzy if-then rules.
Future work includes investigating whether it is possible to improve interpretability in dynamic problems by forgetting past learning.

\vspace{12pt}
%\color{red}
%IEEE conference templates contain guidance text for composing and formatting conference papers. Please ensure that all template text is removed from your conference paper prior to submission to the conference. Failure to remove the template text from your paper may result in your paper not being published.

\end{document}